\documentclass[conference]{IEEEtran}
\IEEEoverridecommandlockouts
\usepackage{graphicx}
\usepackage{booktabs}
\usepackage{multirow}
\usepackage{siunitx}
\usepackage{grffile}
\usepackage{subcaption}
\usepackage{capt-of}
\usepackage{bm}
\usepackage{amssymb}
\usepackage{amsmath}
\usepackage{authblk}
\usepackage{hyperref}
\usepackage{isomath}
\usepackage{paralist}
\usepackage{tabularx}
\usepackage{multicol}

\renewcommand\IEEEkeywordsname{Keywords}

\title{Exploring Test Time Adaptation for Subcortical Segmentation of the Fetal Brain in 3D Ultrasound}

\begin{document}

%
%

\author{Joshua Omolegan$^{1, *}$, Pak Hei Yeung$^{2}$, Madeleine K. Wyburd$^{1}$, Linde Hesse$^{1, \dag}$, Monique Haak$^{3}$ \\
Intergrowth-21\textsuperscript{st} Consortium, Ana I. L. Namburete$^{1}$, Nicola K. Dinsdale$^{1}$
\thanks{$^{*}$Now at Google, $^{\dag}$Now at QuantCo}}



\affil{$^{1}$ Oxford Machine Learning in NeuroImaging Lab, University of Oxford \\ 
$^{2}$ School of Computer Science and Engineering, Nanyang Technological University, Singapore \\
$^{3}$ Department of Obstetrics and Fetal Medicine, Leiden University Medical Center}


%

\maketitle        

\begin{abstract}
Monitoring the growth of subcortical regions of the fetal brain in ultrasound (US) images can help identify the presence of abnormal development. Manually segmenting these regions is a challenging task, but recent work has shown that it can be automated using deep learning. However, applying pretrained models to unseen freehand US volumes often leads to a degradation of performance due to the vast differences in acquisition and alignment. In this work, we first demonstrate that test time adaptation (TTA) can be used to improve model performance in the presence of both real and simulated domain shifts. We further propose a novel TTA method by incorporating a normative atlas as a prior for anatomy. 
In the presence of various types of domain shifts,
we benchmark the performance of different TTA methods and demonstrate the improvements brought by our proposed approach,
which may further facilitate automated monitoring of fetal brain development. 
Our code is available at \url{https://github.com/joshuaomolegan/TTA-for-3D-Fetal-Subcortical-Segmentation}.

\end{abstract}
\vspace*{-5pt}

\section{Introduction}




During pregnancy, ultrasound (US) is routinely used to assess the development of certain brain structures (e.g. subcortical structures), whose volume can serve as biomarkers for neurological conditions \cite{Malinger2020}. However, since these assessments are normally performed in 2D, valuable 3D information is lost, which could provide richer diagnostic information \cite{Hesse2022}.

To utilise 3D shape information, a method to segment subcortical structures is required. However, this is challenging due to the low levels of soft tissue contrast, reverberation artefacts and the presence of speckle. As a result, structural boundaries are often hard to distinguish, causing high levels of inter- and intra-rater variability in manual annotations \cite{Hesse2022}. Furthermore, subcortical segmentation is not a task usually completed in clinical practice, and thus even trained sonographers can have difficulty in accurately segmenting subcortical structures. 

Studies have shown that deep learning (DL) methods can be successfully applied to segmentation tasks in 3D US volumes of the fetal brain \cite{Hesse2022,Wyburd2020cortical}, outperforming traditional image analysis methods. In particular, \cite{Hesse2022} proposed a DL framework for segmenting subcortical regions in the second trimester, achieving promising performance. However, the difficulty of obtaining manual annotations for these structures presents challenges for training DL models. Furthermore, as there is a high variability in possible US acquisition settings, a model trained for a given set of US images may not perform well for images acquired with a different set-up. 
Due to the tendency for DL models to overfit to their training data, changes in image characteristics cause a \textit{domain shift}, normally resulting in a drop in performance and poor quality segmentations. 

\textit{Domain adaptation} methods aim to overcome this domain shift by learning a shared feature representation between source and target domains \cite{Ganin2015,Sun2016}. However, they assume simultaneous and constant access to the source and target data which is normally infeasible: the source data may not be available for the adaptation phase, for instance, due to confidentiality agreements \cite{Bateson2022}, and model retraining is computationally expensive. \textit{Source Free Domain Adaptation} \cite{Bateson2022,Liang2020} relaxes this requirement, enabling the adaptation of models without access to the source data. However, it is assumed that the target data belongs to a single domain: that is, all images are collected with the same acquisition protocol. This assumption is not valid for US data, as the acquisition is often changed for each individual scan to produce the best image. 

To overcome the domain shift commonly seen in US, we explore the use of test-time adaptation (TTA) to adapt an existing subcortical segmentation model \cite{Hesse2022}. TTA adapts models to new data during testing, with access only to the testing data and the pretrained model \cite{Janouskova2024,Wang2021} and can produce a differently adapted model for each test sample or batch. We hypothesise that TTA should be able to overcome the large diversity of domain shifts between US scans. Our contributions are as follows:
\begin{compactitem} 
    \item We demonstrate that TTA can be used to adapt models to US specific domain shifts, without needing additional manual segmentation, and explore the use of novel adaptations to the existing Test Entropy Minimisation (TENT) method \cite{Wang2021} to make TTA more suitable for fetal US.
    \item We propose a novel TTA method (EntropyKL) to incorporate an US atlas \cite{Namburete2023} as a prior for the expected volume of the each subcortical region respectively, and show how its incorporation increases the performance of model adaptation across a range of domain shifts.
    \item We benchmark the performance of a range of TTA approaches in the presence of simulated domain shifts, real domain shifts across vendors, and domain shift due to varying gestational week, and demonstrate the improved performance of EntropyKL. 
\end{compactitem}

\section{Method}

\begin{figure}[htbp]
    \centering
    \includegraphics[width=0.45\textwidth]{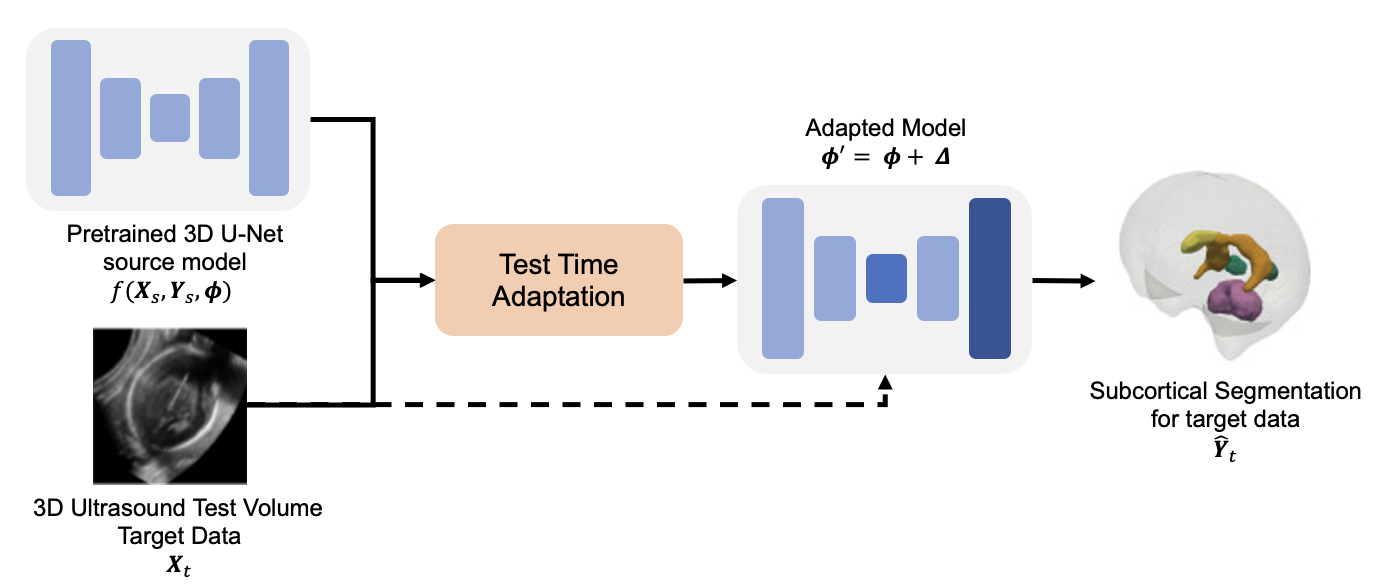}
    \caption{TTA framework schematic. The source model is the pretrained 3D UNet from \cite{Hesse2022}, which is adapted on the target data and then used on the same data to make a prediction. 
    }
    \label{fig:TTA schematic}
    \vspace*{-5pt}
\end{figure}

We assume access to a pretrained \textit{source model}, $f(\bm{X}_s, \bm{Y}_s; \bm{\phi})$, where $\bm{\phi}$ are learned weights and \{$\bm{X}_s, \bm{Y}_s\}$ are the labelled training data pairs from the dataset $D_s$. We wish to adapt the model, $f$, as to achieve the maximum performance for each sample in a target data $D_t = \{\bm{X}_t\}$ for which no labels are available. A schematic demonstrating the approach can be seen in Fig. \ref{fig:TTA schematic}. 
In this section, we first introduce the adaptation of different TTA approaches to fetal US subcortical segmentation. 
Then, we describe our proposed TTA technique to incorporate an US atlas \cite{Namburete2023} as a prior for the expected volume of the subcortical regions.

\subsection{Test Entropy Minimisation}
We use Test Entropy Minimisation (TENT) \cite{Wang2021} as our baseline TTA approach, as it has shown success with medical images \cite{valanarasu2024fly}. The goal of TENT is to minimise the entropy, $H(\bm{\hat{Y}})$, of the model predictions, where $\bm{\hat{Y}} = f(\bm{X}_t, \bm{\phi}')$ and ${\bm{\phi}'}$ are the adapted model weights ${\bm{\phi}} + \Delta$, where $\Delta$ is a small change, based on the observation that model predictions with the highest entropy are the most likely to be incorrect \cite{Wang2021}. Thus, TENT minimises the Shannon entropy:
\begin{equation}
    L_{TENT} = - \sum_c p(\bm{\hat{Y}}_c)log(p(\bm{\hat{Y}}_c))
\end{equation}
\vspace*{-1pt}

\noindent where $p(\bm{\hat{Y}}_c)$ is the softmax output from $f$ for class $c \in C$ and $C$ is the total number of segmentation classes.

However, na\"ive minimisation of this loss would lead to the model moving too far away from the source learned feature parameters $\bm{\phi}$ and thus the loss of the source model knowledge. Therefore, TENT only optimises the weights of the batch normalisation layers as these are linear and low-dimensional, constraining the model weight updates and reducing the likelihood of collapsing to a trivial solution. 

\subsection{LayerInspect}
Modifying the parameters of only the batch normalisation layers potentially limits the types of distribution shift we can adapt to. Thus, to allow for more flexible adaptation, we consider updating layers other than the batch normalisation layers. As we still need to not deviate too far from the original model, inspired by pruning methods \cite{Dinsdale2022}, we propose a new TTA method, LayerInspect. In this, we select the $m$ layers with the largest difference in magnitude between the source and target activations to be updated. The source activations need only be calculated once (at the end of model training), so the source data does not need to be stored.

Following the notation in \cite{Dinsdale2022}, consider the feature activation maps in a network, $f$, denoted by $\bm{z}_l^{(k)}$ where $ l \in {1... L}$ is current layer in the model $f$ and $k$ refers to the $k^{th}$ filter in layer $l$. To assess the magnitude of the activations, we calculate the Taylor approximation proposed in \cite{Molchanov2017}:
\begin{equation}
    \bm{\theta}_{TE} (\bm{z}_l^{(k)}) = \left| \frac{1}{N} \sum_n \frac{\delta L_{TENT}}{\delta \bm{z}_{l,n}^{(k)}}  \bm{z}_{l,n}^{(k)} \right|
\end{equation}
where  N is the total number of data points. We utilise the Taylor approximation as it accounts for gradient information that may be useful for determining the layers with the largest impact on the models output. The activations are then normalised using L2 normalisation, to account for differences in scale with model depth \cite{Molchanov2017}, and finally the difference is found as $|\bm{\theta}_{TE}^s - \bm{\theta}_{TE}^t|$.

\subsection{Incorporating Atlas Prior via EntropyKL}
Finally, following \cite{Bateson2022}, we considered the incorporation of a class ratio prior to the loss function, providing the model with an estimate of the amount of each tissue that should be present, preventing the model from collapsing to the trivial solution. In \cite{Namburete2023}, the authors used fetal brain ultrasound scans collected as part of the INTERGROWTH-21st Project \cite{papageorghiou2018intergrowth} to create a normative atlas of the average healthy fetal brain at various weeks of pregnancy. Additionally, the authors generated segmentation maps of the atlas by manual segmentation. We use this as our estimate of the expected amount of each tissue, and integrate it into our TTA methods by combining it with the Entropy minimization objective from TENT. Our new loss function is:
\begin{equation}
    L_{ENTROPYKL} = L_{TENT} + \lambda KL(\hat{\tau}_t || \tau_t)
\end{equation}
where $\operatorname{KL}(p || q) = p \log \left( \frac{p}{q} \right)$ is the KL divergence - a measure of the difference between two probability distributions, $\tau(k)$ is the proportion of the atlas that was class $k$, $\hat{\tau}(k)$ is the average probability of a pixel in the image belonging to class $k$, and $\lambda$ is a hyperparameter chosen to weight the contribution of the KL divergence term. This prior gives our model some idea of how much of each class should be present and penalises updates to the model parameters that drastically change the ratio of classes. Note that we still only update the batch norm parameters using this new objective for the same reasons as in TENT.

If the image we wish to adapt our model to is not a healthy fetal brain (\textit{e.g.} if a disease or birth defect caused the domain shift), the ratio of classes may not be similar to the ratio of classes in our atlas. To address this, we use the hyperparameter $\lambda$, which controls the weighting of the KL divergence compared to the entropy part of the loss function. Low values of $\lambda$ reduce our loss function to be identical to TENT, which may be favorable if our volumes differ largely from the atlas, while high values of $\lambda$ will essentially force the model to predict the exact ratio of classes seen in the atlas. We explored the effect of the value of $\lambda$ (see Figure \ref{fig:layer-inspect-hyperparam}).

\section{Experimental Setup}

\noindent \textbf{Source Model:} We used the model from \cite{Hesse2022} as the source model, which used few shot learning to segment four subcortical regions: choroid plexus (CP), lateral posterior ventricle horn (LPVH), cerebellum (CB) and cavum septum pellucidum et vergae (CSPV) during the second trimester. The model was trained from manual annotations on 3D US images acquired by the INTERGROWTH-21st Fetal Growth Longitudinal Study \cite{papageorghiou2018intergrowth}, collected using a Philips HD9 curvilinear probe. 

\begin{figure}
    \centering
    \begin{subfigure}{0.45\textwidth}
        \centering
        \includegraphics[width=\textwidth]{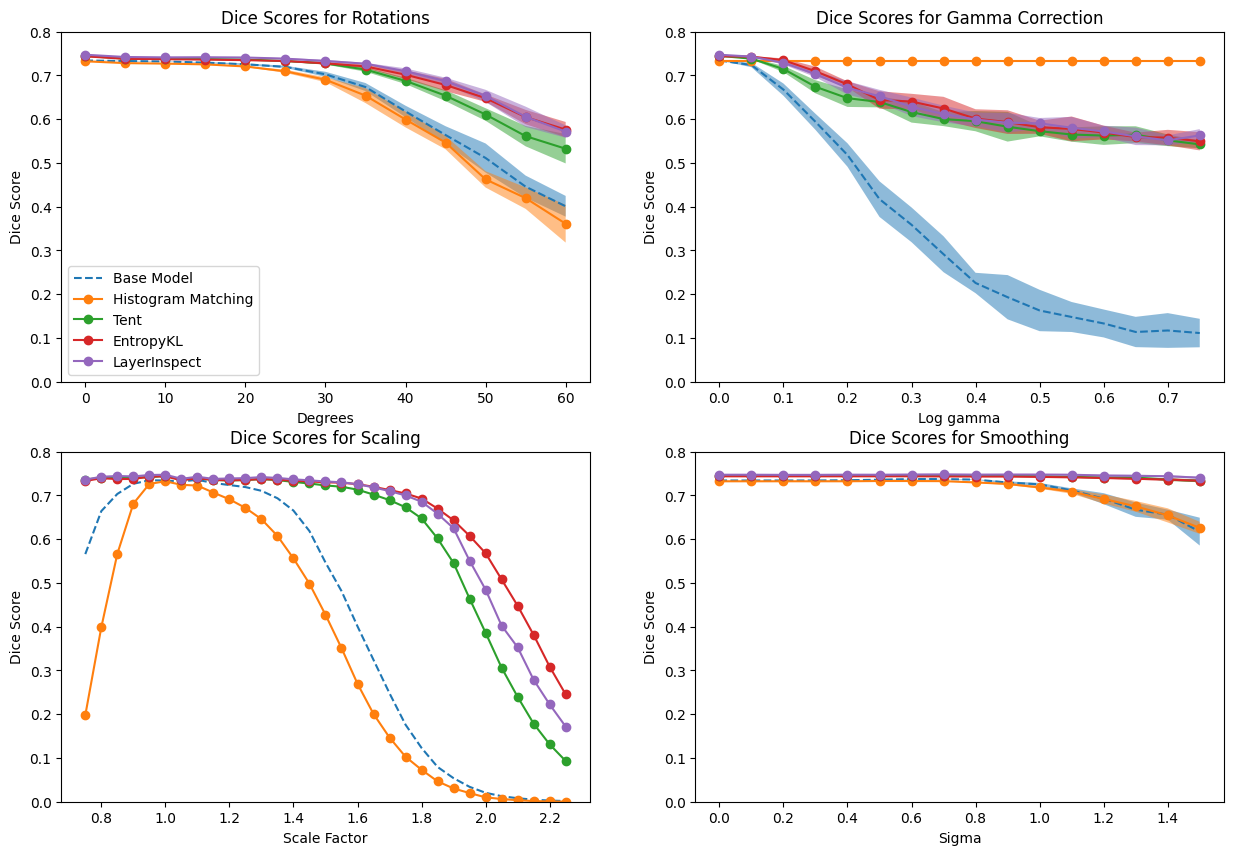}
        \caption{Full dataset adaptation}
        \label{fig:fulldataset}
    \end{subfigure}
    \vfill
    \begin{subfigure}{0.45\textwidth}
        \centering
        \includegraphics[width=\textwidth]{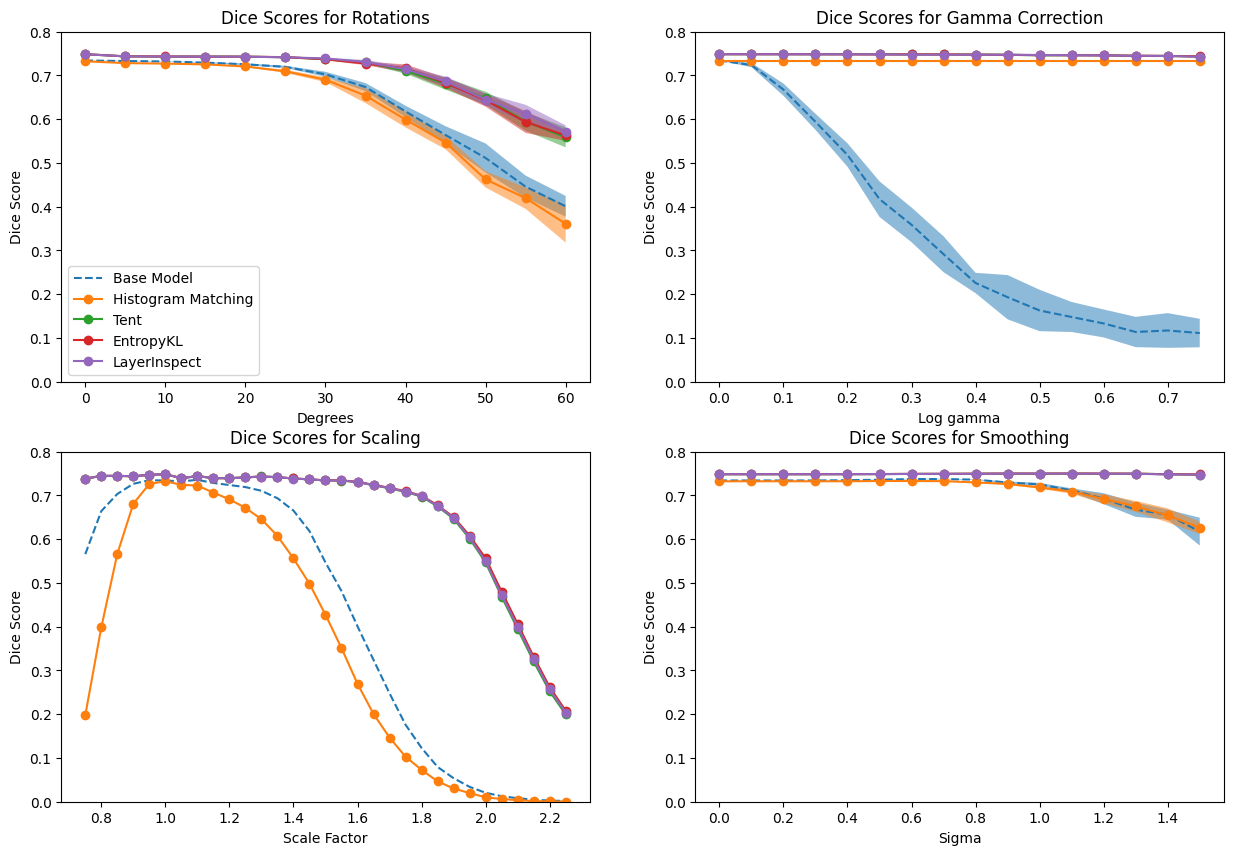}
        \caption{Single Sample Adaptation}
        \label{fig:singlesample}
    \end{subfigure}
    \caption{Results from adapting to simulated augmentations. Dice scores are averaged across all four subcortical regions, and the shaded region indicates the standard deviation bounds.}
    \label{fig:simulated shifts}
    \vspace*{-15pt}
\end{figure}

\noindent \textbf{Simulated Domain Shifts:} We first considered a range of simulated domain shifts, designed to capture expected variation between US scans, going beyond the magnitude of the augmentations used in \cite{Hesse2022}. We used 59 pre-aligned volumes at 21 gestational weeks from the INTERGROWTH-21st study \cite{papageorghiou2018intergrowth}, with propagated annotations from an atlas \cite{Hesse2022,Namburete2023}. All images are of size $160\times160\times160$ voxels, with an isotropic voxel size of 0.6mm. The considered domain shifts were: rotations, scaling, Gaussian smoothing and contrast changes (Gamma Correction), implemented using TorchIO \cite{perez-garcia_torchio_2021}. These simulated shifts and propagated annotations allow us to quantitatively evaluate the performance of the TTA approaches.

\begin{figure}
    \centering
     \includegraphics[width=0.2\textwidth]{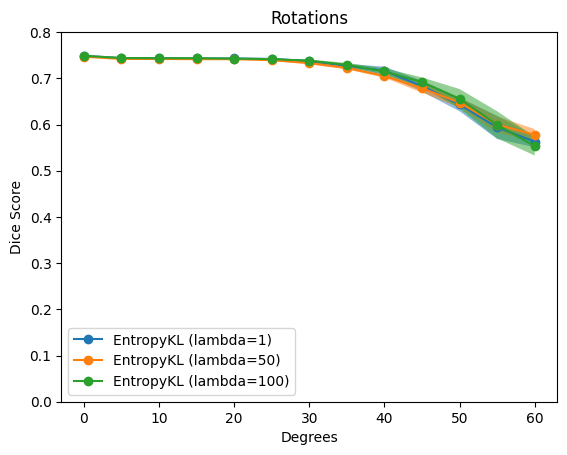}
     \includegraphics[width=0.2\textwidth]{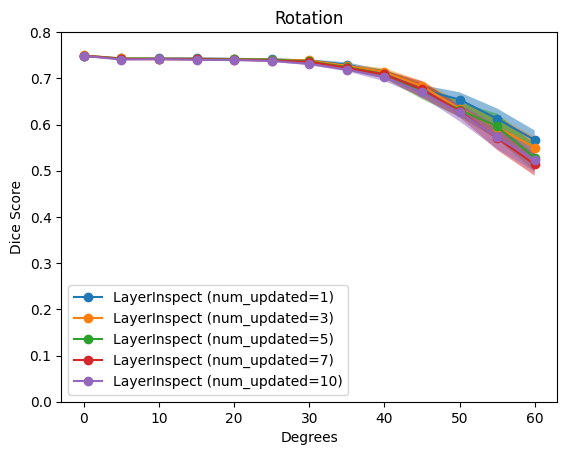}\includegraphics[width=0.2\textwidth]{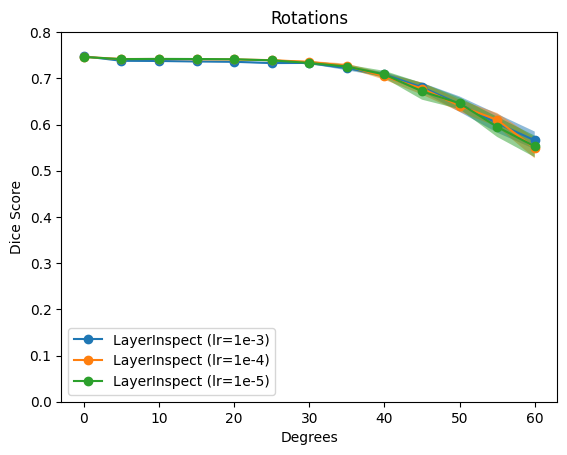}
    \caption{EntropyKL with varying $\lambda$ (a). LayerInspect performance for various rotations, varying the number of layers updated (b) and the learning rate (c).}
    \label{fig:layer-inspect-hyperparam}
    \vspace*{-15pt}
\end{figure}

\textbf{Augmentation Details}: Rotations and scaling are both performed along each dimension. Given a value $x$, each image is rotated by a value randomly selected from the range $[-x, x]$. For gamma correction, this value of $\log \gamma$ is also sampled from $[-x, x]$. The value of $\sigma$ for Gaussian smoothing is sampled from $[0, x]$. Note that augmentations were used during the training of our base model, and so we chose ranges that exceed the range of augmentation seen during training, but are still realistic. This ensures that the data is out-of-distribution. 

\noindent \textbf{Gestational Age:}
The source model was trained on fetal images acquired between $18^{+0 \mathrm{days}}$ and $26^{+6 \mathrm{days}}$ gestational weeks (GW). However, the appearance and size of the subcortical structures we consider in this work develop drastically over this period \cite{Namburete2023}, and a single model may be suboptimal across this variation. Thus, we consider using TTA to adapt the model across this age range, considering the performance separately across each week. This gave a dataset of 529 labelled images with atlas-propagated annotations. 

\noindent \textbf{Unseen Datasets:}
Finally, we consider data from two unseen scanners (Canon and GE) collected at the Leiden University Medical Centre \cite{yeung2024implicitvol}. We again consider scans from 21 GW and both datasets have been preprocessed in an identical way to the INTERGROWTH-21st study, and so have been aligned, are of size 160×160×160 voxels, and have an isotropic voxel size of 0.6mm. This resulted in 31 volumes from the Canon dataset and 22 from the GE dataset. Manual annotations are not available so only qualitative analysis is completed. 

\noindent \textbf{Implementation:} For each experiment, TTA was performed on an Nvidia A10 GPU. The Adam optimiser was used with a learning rate of $10^{-3}$ for both TENT and EntropyKL, and a learning rate of $10^{-4}$ for LayerInspect. Due to memory constraints, a batch size of 2 was used for all methods, and the number of backwards passes used for model adaptation was the smallest value for which we saw a substantial change (Unseen datasets - 25, 1 for all others). We believe more backward passes were required for the unseen datasets because of the degree of domain shift present, as these shifts are likely equivalent to many different individual augmentations. The value of $\lambda$ for EntropyKL was set to 1 and for LayerInspect only $m=1$ layers were updated (see Figure \ref{fig:layer-inspect-hyperparam} for sensitivity experiments). 

\begin{figure}
    \centering
    \includegraphics[width=0.48\textwidth]{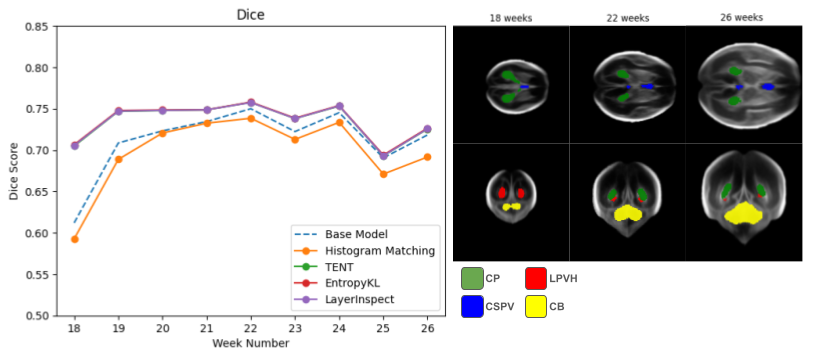}
    \caption{LHS: Average Dice score across all structures at each gestational age. Differences between methods are negligible and so overlap on the graph.  RHS: The same plane from the US atlas \cite{Namburete2023} at different ages, showing the development of the four subcortical structures.}
    \label{fig:gestational week}
    \vspace*{-15pt}
\end{figure} 

\section{Results}
 \textbf{Simulated Domain Shifts:} We test both adapting to the full dataset and to a single volume at a time. Although it may be reasonable in many domains to assume all data from a site is a single domain \cite{Dinsdale2021Harm}, the flexible nature of US scan acquisition means this is unlikely. However, as TENT assumes access to batches of data \cite{Wang2021}, we compare both approaches. We also present two baselines: the original source model and histogram matching, as standard preprocessing approach. We compare Dice scores, averaged across the subcortical regions. 

 Fig. \ref{fig:simulated shifts} shows the results. All TTA methods provided significant improvement (paired t-test, $p<0.05$) over the base model, indicating their effectiveness in overcoming the simulated domain shifts. They each have similar performance, but the proposed adaptations to TENT in LayerInspect and EntropyKL lead to improvement, with EntropyKL being the most robust to large transformations. Histogram matching provides little advantage over the base model apart from Gamma correction, where it effectively removes the applied augmentation. 

Comparing the results, we see TTA performs better when adapting to a single volume than to a batch across the range of transformations. This improvement, and the nature of each US image effectively being its own imaging domain due to the flexible nature of US imaging, motivates the use of single volume adaptation going forward.

\noindent \textbf{Gestational Age:} Figure \ref{fig:gestational week} compares our baselines with the TTA methods across gestational age. Considering three ages from the atlas, we see the structures change significantly throughout gestation, especially in the volume of the brain they occupy. DL models are prone to converging to the average \cite{Dinsdale2021}, and it can clearly be seen that the source model performs worse on the early GW (average Dice = 0.612) where the structures are the most different. Applying TTA methods significantly increases the performance of the model for these weeks, and never reduces performance. It is worth noting that histogram matching reduces performance because the transformations are not intensity based but morphometric, but for other methods the model learned to overcome these changes.

\begin{figure}[t]
    \centering
    \includegraphics[width=0.48\textwidth]{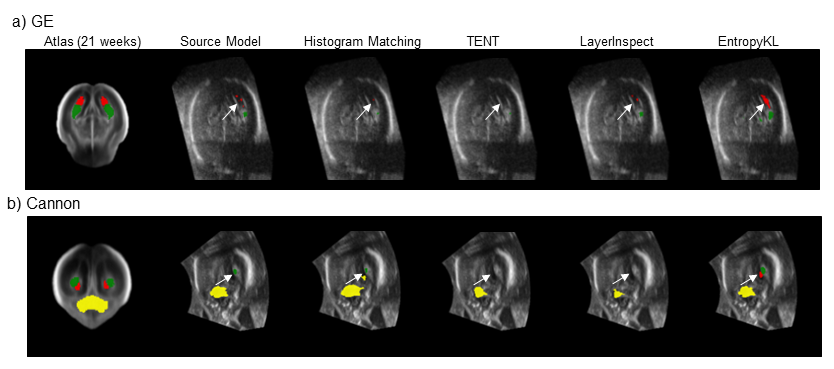}
    \caption{An example slice from a scan from a) a GE scanner and b) a Canon scanner, comparing the performance of the different methods, with the reference atlas slice to show expected anatomy. Red=LPVH, Green=CP, Yellow=CB.}
    \label{fig:liden}
    \vspace*{-15pt}
\end{figure}

\noindent \textbf{Unseen Datasets:} Finally, we consider two unseen datasets where no ground truth labels are available. As mentioned previously, we had to increase the number of backward passes, indicating the large degree of domain shift. Fig. \ref{fig:liden} shows example segmentations from the source model, TTA methods, and the reference atlas. Labels are only predicted for the distal hemisphere, and with no manual labels we can only make qualitative comparison. We show a sample from each scanner manufacturer where the performance of the base model was the most uncertain (highest entropy), and so we expect the greatest benefit from the TTA. For both samples, it can clearly be seen that the addition of the atlas prior (EntropyKL) leads to the segmentations that best represent the expected anatomy and segmentation of all of the subcortical regions present in the slice, showing the advantage of incorporating prior knowledge. The prior only provides information about the expected ratio of tissues and not location: thus, the model is not simply predicting the atlas but learning a more informed update.

\section{Discussion}
Our results demonstrate that TTA can adapt subcortical segmentation models to domain shifts in US scans, showing improvements of up to 0.6 in Dice score for synthetic augmentations, and 0.1 for gestational age. Given the challenges of manually segmenting these regions in US images this may be key for the identification of abnormal fetal development. 

We further proposed a novel TTA technique to incorporate the normative altas \cite{Namburete2023} as a prior for expected volume, which improved performance over the other TTA methods, especially for the scans from unseen sites. However, since the atlas corresponds to healthy fetal growth, it may bias the model incorrectly in the presence of fetal abnormality. Future work will explore its use in this case.  

\section{Acknowledgments}

PH. Yeung is funded by the Presidential Postdoctoral Fellowship from the Nanyang Technological University. A.I.L. Namburete, M.K.Wyburd and N.K. Dinsdale are grateful for support from the Bill and Melinda Gates Foundation.

\bibliographystyle{splncs04}
\bibliography{references}

\end{document}